%% file: acl_latex.tex
\pdfoutput=1

\documentclass[11pt]{article}

\usepackage[]{acl}

\usepackage{times}
\usepackage{latexsym}

\usepackage[T1]{fontenc}

\usepackage[utf8]{inputenc}

\usepackage{microtype}
\usepackage{booktabs}
\usepackage{multirow}
\usepackage{graphicx}
\usepackage{xcolor}

\definecolor{betterGreen}{RGB}{60,210,50}

\usepackage{subfiles} 
\title{ScAN: Suicide Attempt and Ideation Events Dataset}

 \author{Bhanu Pratap Singh Rawat\textsuperscript{1,*}, Samuel Kovaly\textsuperscript{1,\#}, Wilfred R. Pigeon\textsuperscript{2,3,$\dagger$}, Hong Yu\textsuperscript{1,3,4,$\ddagger$}\\
  \textsuperscript{1}CICS, UMass-Amherst, \textsuperscript{2}University of Rochester, \textsuperscript{3}U.S. Department of Veterans Affairs, \\ \textsuperscript{4}Center of Biomedical and Health Research in Data Sciences \\
  \texttt{\textsuperscript{*}brawat@umass.edu, \textsuperscript{\#}skovaly@umass.edu,} \\ \texttt{ \textsuperscript{$\dagger$}wilfred\_pigeon@urmc.rochester.edu, \textsuperscript{$\ddagger$}hong\_yu@uml.edu} \\} 

\begin{document}
\maketitle
\begin{abstract}

Suicide is an important public health concern and one of the leading causes of death worldwide. Suicidal behaviors, including suicide attempts (SA) and suicide ideations (SI), are leading risk factors for death by suicide.
Information related to patients' previous and current SA and SI are frequently documented in the electronic health record (EHR) notes.
Accurate detection of such documentation may help improve surveillance and predictions of patients' suicidal behaviors and alert medical professionals for suicide prevention efforts.
In this study, we first built
\textbf{S}ui\textbf{c}ide \textbf{A}ttempt and Ideatio\textbf{n} Events (ScAN) dataset, a subset of the publicly available MIMIC III dataset spanning over $12k+$ EHR notes with $19k+$ annotated SA and SI events information. The annotations also contain attributes 
such as method of suicide attempt.
We also provide a strong baseline model ScANER (\textbf{S}ui\textbf{c}ide \textbf{A}ttempt and Ideatio\textbf{n} \textbf{E}vents \textbf{R}etreiver), a multi-task Ro\textsc{BERT}a-based model with a \emph{retrieval module} to extract all the relevant suicidal behavioral evidences from EHR notes of an hospital-stay and, and a \emph{prediction module} to identify the type of suicidal behavior (SA and SI) concluded during the patient's stay at the hospital. 
ScANER 
achieved a macro-weighted F1-score of $0.83$ for identifying suicidal behavioral evidences and 
a macro F1-score of $0.78$ and 
$0.60$ for classification of SA and SI for the patient's hospital-stay, respectively. ScAN and ScANER are publicly available\footnote{The annotations, code and the models are availble at \href{https://github.com/bsinghpratap/ScAN}{https://github.com/bsinghpratap/ScAN}.}. 
\end{abstract}


\section{Introduction}
\subfile{sections/intro}

\section{Related Works}
\subfile{sections/related_works}
\section{Dataset}
\subfile{sections/dataset}
\section{Methodology}
\subfile{sections/methodology}

\section{Results and Discussion}
\subfile{sections/results}

\section{Conclusion}
In this paper, we introduce ScAN: a publicly available suicide attempt (SA) and ideation (SI) events dataset that consists of $12,759$ EHR notes with $19,960$ unique evidence annotations for suicidal behavior. To our knowledge, this is the largest and publicly available dataset for SA and SI, an important resource for suicidal behaviors research. 
We also provide a strong Ro\textsc{bert}a baseline model for the dataset: ScANER (SA and SI retriever) which consists of two sub-modules: (a.) an \textit{evidence retriever module} that extracts all the relevant evidence paragraphs from the patient's notes and (b.) a \textit{prediction module} that evaluates the extracted evidence paragraphs and predicts the SA and SI event label for the patient's stay at the hospital. ScAN and ScANER could help extract suicidal behavior in patients for suicide surveillance and predictions, leading to potentially early intervention and prevention efforts by medical professionals.

\bibliography{anthology,custom}
\bibliographystyle{acl_natbib}
\onecolumn
\appendix
\subfile{sections/appendix}

\end{document}

%% file: sections/intro.tex
For decades, suicide has been one of the leading causes of death 
\cite{suicide_survey}. 
The suicide rate in the United States increased from $10.5$ per $100,000$ in $1999$ to $14.2$ in $2018$, a $35\%$ increase \cite{hedegaard2020increase}. Globally, $740,000$ people commit suicide each year. The rates of suicidal behaviors, suicide attempt (SA) and suicide ideation (SI), are much higher  
\cite{who2022fact}.
\begin{figure}[!t]
    \centering
    \includegraphics[width=7.7cm]{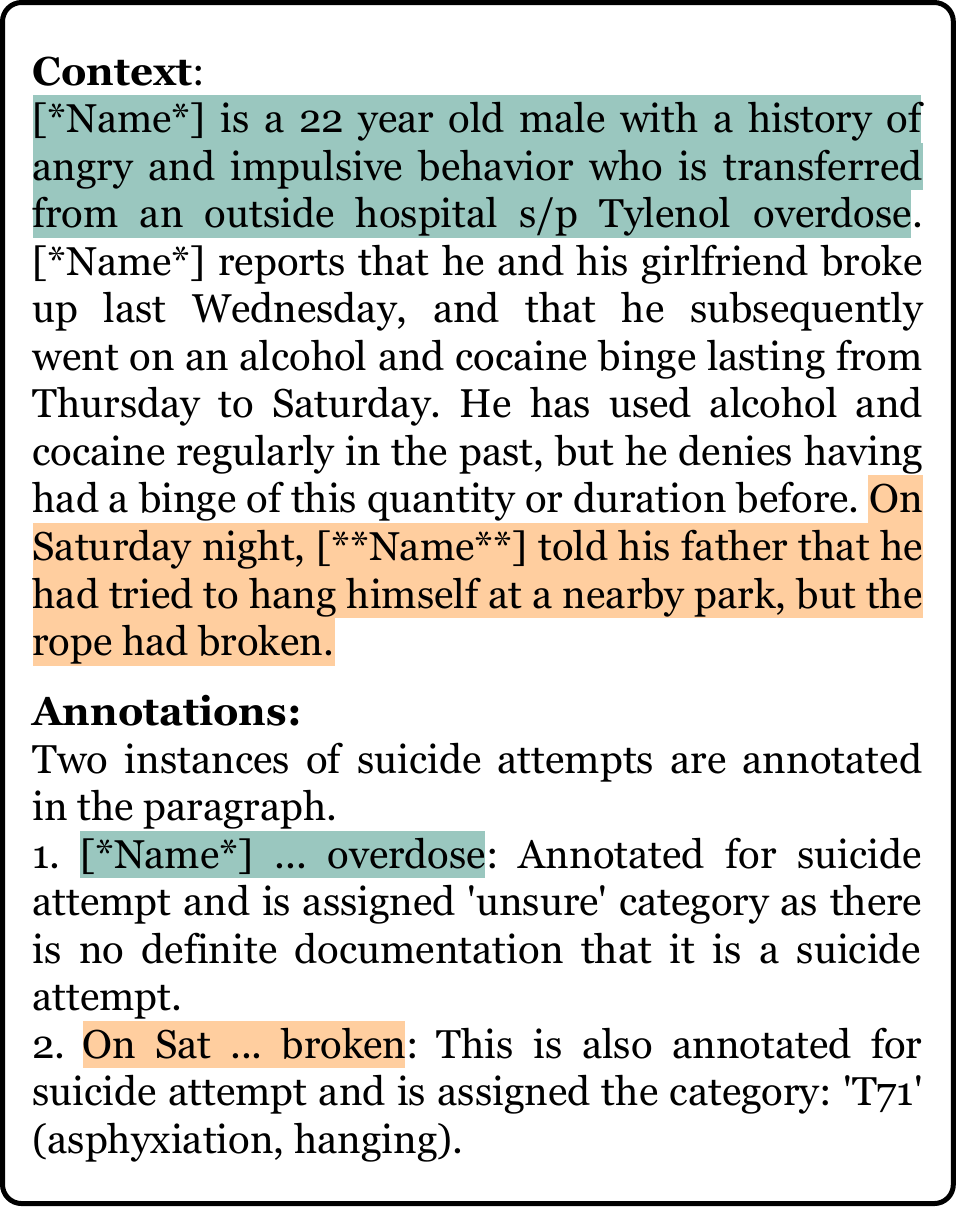}
    \caption{An example of \textit{positive} and \textit{unsure} evidence annotations for SA in an EHR note.}
    \label{fig:annotation}
    \vspace{-1.5em}
\end{figure}

A prior study shows that a large proportion of suicide victims sought care well before their death \cite{kessler2020suicide}. Suicidal behaviors, including SA and SI are recorded by clinicians in electronic health records (EHRs).  
This knowledge can in turn help clinicians assess risk of suicide and make prevention efforts \cite{jensen2012mining}.  
The diagnostic ICD codes include suicidality codes for both SA and SI. However a study has shown that ICD codes can only capture 3\% SI events, while 97\% of SI events are described in notes \cite{anderson2015monitoring}. 
In addition, of patients described with SA in their EHR notes, only 19\% had the corresponding ICD codes \cite{anderson2015monitoring}. 
Therefore, it is important to develop natural language processing (NLP) approaches to capture such important suicidality information. 


Researchers have developed NLP approaches to detect SA and SI from EHR notes \cite{metzger2017use, downs2017detection, fernandes2018identifying, cusick2021using}. These studies either used rule-based approaches \cite{downs2017detection, fernandes2018identifying, cusick2021using} or built the SA and SI identification models on a small set \cite{metzger2017use} or private set \cite{bhat2017predicting, tran2013integrated, haerian2012methods} of EHR notes.
It is also difficult to compare the results of those studies as they varied in EHR data, data curation, as well as NLP models, which were not made available to the public. 

In this study, we present ScAN: \textbf{S}ui\textbf{c}ide \textbf{A}ttempt and Ideatio\textbf{n} Events Dataset, a publicly available EHR dataset that is a subset of the MIMIC III data \cite{johnson2016mimic}. ScAN contains $19,690$ expert-annotated SA and SI events
with their attributes (e.g., methods for SA) over $12,759$ EHR notes. Specifically, experts annotated suicidality evidence or sentences relevant to SA and SI events during a patient's stay at the healthcare facility, an example of SA annotations is shown in Fig~\ref{fig:annotation}. The evidences were put together to assess whether the patient has an SA or SI event.   

We also present ScANER (\textbf{S}ui\textbf{c}ide \textbf{A}ttempt and Ideatio\textbf{n} \textbf{E}vents \textbf{R}etriever), a Ro\textsc{BERT}a-based NLP model that is built on a multi-task learning framework for retrieving evidences from the EHRs and then predicting a patient's SA or SI event using the complete set of EHR notes from the hospital stay using a multi-head attention model.
We focus on the prediction of SA and SI using all the EHR notes during a patient's stay because for the whole duration, multiple EHR notes and note types are generated, including \emph{admission} notes, \emph{nursing} notes, and \emph{discharge summary} notes. Suicidal information are described in multiple notes throughout the stay. For example, a patient was admitted to the hospital with opioid overdose. It was documented initially in the admission note as an SA, but later dismissed as an accident after physician's evaluation. In another example, an opioid overdose admission was first documented as an accident on admission, but later documented to be an SA event after clinical assessment. Both ScAN and ScANER capture SA and SI information at the hospital-stay level. 
ScANER is able to retrieve suicidal evidences from EHR notes with a macro-weighted F1-score of $0.83$ and is able to predict SA and SI with a macro F1-score of $0.78$ and $0.60$, respectively. Our annotation guidelines, ScAN, and ScANER system will be made publicly available, making ScAN a benchmark EHR dataset for SA and SI events detection. We will release the training and evaluations splits used in this study for benchmarking new models.

%% file: sections/related_works.tex

Efforts on detecting SA and SI within EHRs have been explored in recent years. Most work used rule-based or traditional machine learning-based approaches. In one study, experts created handcrafted rules from mentions of suicidality (both SA and SI) and then used the rules to identify suicidality as positive, negative, or unknown in a document \cite{downs2017detection}. The rule-based approaches are limited by their scalability.
In another study, structured and unstructured EHRs were used to classify at the hospital-stay level as SA, SI, or no mention of suicidal behavior \cite{metzger2017use}.  
The training data consisted of only 112 SA, 49 SI and 322 unrelated examples. In contrast, ScAN comprises of 697 hospital-stays with more than 19,000 suicidal event examples over 12,759 clinical notes. Only traditional machine learning models such as random forest \cite{breiman2001random} were explored. In contrast, ScANER was built on the state-of-the-art self-attention based model.

   

Hybrid approaches have also been developed to identify SA at the hospital-stay level \cite{fernandes2018identifying}. 
In that study, a post-processing heuristic rule-based filter (e.g., removing negated events) was applied to the machine-learning-based classifier (a SVM \cite{cortes1995support} classifier) to reduce false positives. Training and evaluation were done also on relatively small datasets (500 for training and 500 for evaluation).

Finally, weakly supervised approaches have been developed to identify SI from EHRs \cite{cusick2021using}. In that study, authors used ICD codes to identify 200 patients with SI and then obtained EHR notes of those patients ($6,588$). This EHR note dataset was then used as the `current' SI training data. The remaining 400 patients were labelled as `potential' SI and their $12,227$ EHRs were also labelled the same. Authors used multiple statistical machine learning models and one deep learning model:  convolutional neural network. \cite{bhat2017predicting} also used feed-forward neural networks to predict suicide attempts over 500k unique patients but the EHR data for this study is not publicly available. \cite{ji2020suicidal} surveyed multiple studies where the researchers worked on private datasets \cite{tran2013integrated, haerian2012methods} for suicide attempt and ideation prediction. Whereas in our study, in contrast to using the ICD codes which has considerable errors, domain experts chart-reviewed a large, publicly available set of EHRs for SI and SA,
along with their attributes (e.g., positive or negative SA, SI and the type of self-harm such as asphyxiation and overdose). 

%% file: sections/dataset.tex
In this section, we introduce ScAN (\textbf{S}ui\textbf{c}ide \textbf{A}ttempt and Ideatio\textbf{n} Events Dataset) and describe it's data collection and annotation process. We also discuss some examples from ScAN along with basic dataset statistics.

\subsection{Dataset collection}
For annotation, we selected the notes from the MIMIC-III \cite{johnson2016mimic} dataset, which consists of the de-identified EHR data of patients admitted to the Beth Israel Deaconess Medical Center in Boston, Massachusetts from 2001 to 2012 \cite{johnson2016mimic}. The data includes notes, diagnostic codes, medical history, demographics, lab measurements among many other record types. We chose MIMIC-III because it is publicly available under a data use agreement and allows clinical studies to be easily reproduced and compared. 

The diagnostic ICD codes for the patients are provided at hospital-stay level in MIMIC with admission identification numbers (HADM\_ID in MIMIC database). We first filtered the hospital stays that had ICD codes associated with suicide and overdose. This resulted in $697$ hospital-stays for $669$ unique patients. For each stay, multiple de-identified notes such as nursing notes, physician notes, and discharge summaries are available. For the selected $697$ hospital-stays we extracted a total of $12,759$ notes. Each medical note contains multiple sections about a patient such as family and medical history, assessment and plan, and discharge instructions. We extracted different sections from these clinical notes using MedSpaCy’s\footnote{https://github.com/medspacy/medspacy} \texttt{clinical\_sectionizer} and filtered the relevant sections from these clinical notes for annotation. The extensive list of these sections is provided in Appendix~\ref{app:clinical_sections}.

\subsection{Annotation Process}
The aim was to annotate all instances of SA and SI documented in the medical notes as defined by Center of Disease Control and Prevention (CDC) \cite{hedegaard2020increase}. The filtered $12,759$ notes were annotated by a trained annotator under the supervision of a senior physician. Each note consisted of instances of SA, SI, both or none. The senior physician randomly annotated $330$ notes and had a 100\% agreement with the annotator on hospital-stay level annotation and $85$\% agreement on sentence-level annotations. After adjudication between the senior physician and the annotator, the disagreements were discussed and adjusted by the annotator.

\begin{figure}[!htbp]
    \centering
    \includegraphics[width=7.7cm]{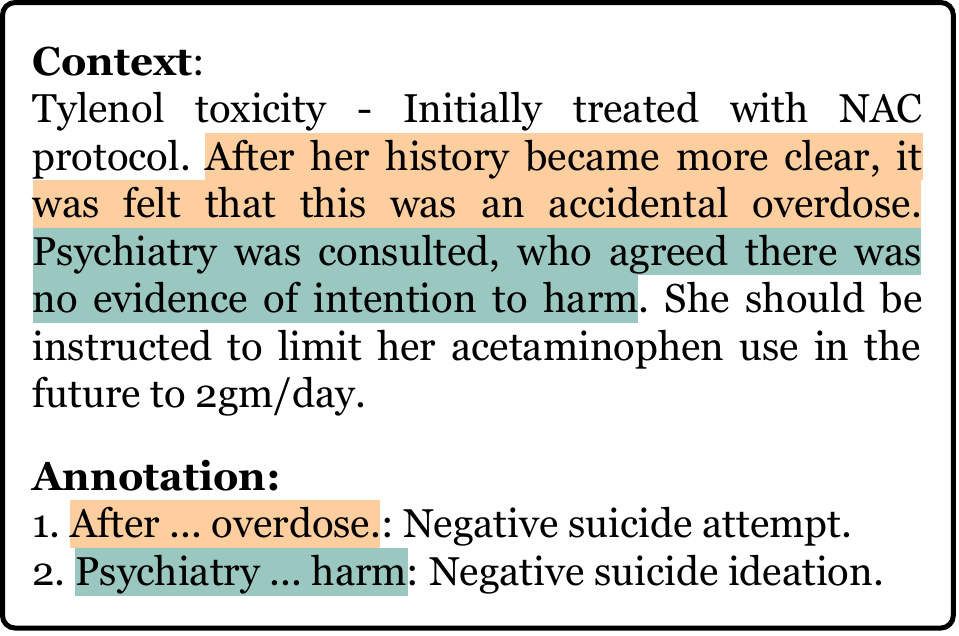}
    \caption{An example with \textit{negative} SA and \textit{negative} SI annotations.}
    \label{fig:neg_sa}
    \vspace{-1.5em}
\end{figure}

\paragraph{Suicide Attempt (SA):}The annotator labelled all the sentences with a mention of SA. 
Some hospital stays could represent multiple types of SA, such as in Fig.~\ref{fig:annotation}, where `tried to hang himself' is labelled as a \textit{positive} SA and `Tylenol overdose' is labelled as \textit{unsure} since the overdose was never confirmed as an SA event elsewhere in the medical notes of the patient's hospital-stay. The label \textit{unsure} is used when it is not clearly documented if a self-harm was an SA event or not. The \textit{negative} instance, example shown in Fig.~\ref{fig:neg_sa}, is a sentence that confirms that the self-harm, an ``accidental overdose'', responsible for the patient's hospital-stay is not an SA event. In this work, we only focused on suicidal self-harm and not non-suicidal self-harm \cite{crosby2011self}.

Further sub-categories are also provided for an SA annotation in the form of the ICD label group: a.) T36-T50: Poisoning by drugs, medications and biological substances  b.) T51-T65: Toxic effects on non-medical substances c.) T71: Asphyxiation or suffocation and d.) X71-X83: Drowning, firearm, explosive material, jumping from a high place, crashing motor vehicles, other specified means. 

\begin{figure}[!htbp]
    \centering
    \includegraphics[width=7.7cm]{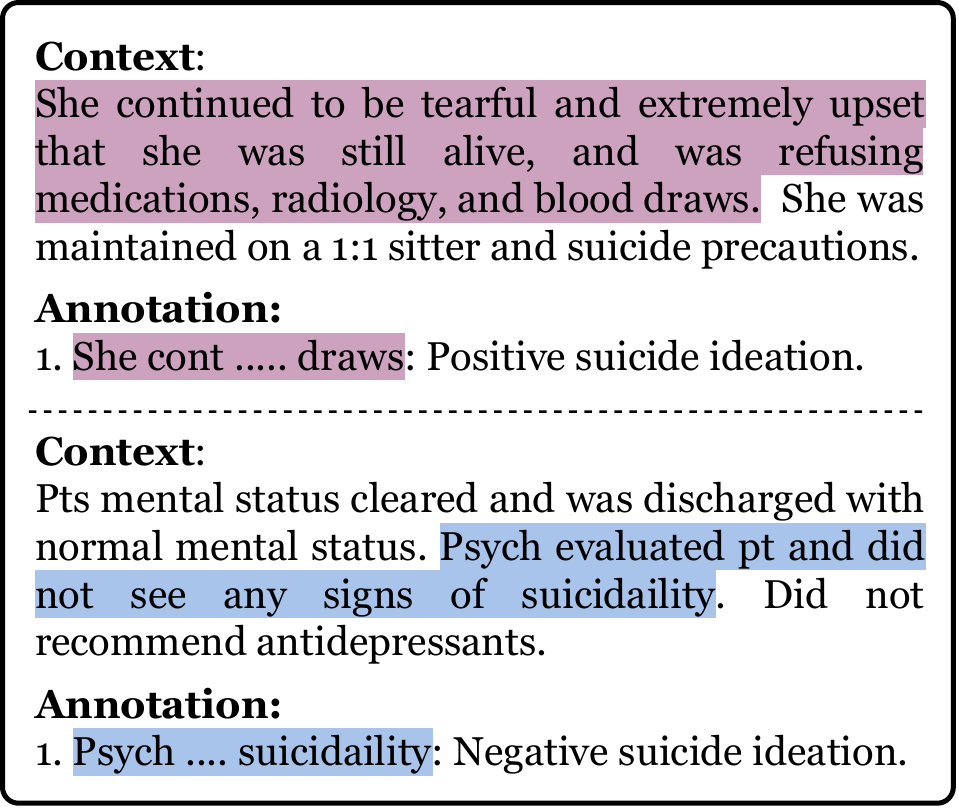}
    \caption{Examples of \textit{positive} and \textit{negative} SI annotations.}
    \label{fig:pos_neg_si}
    \vspace{-0.5em}
\end{figure}

\paragraph{Suicide Ideation (SI):}SI is defined as any mention and/or indication of wanting to take one’s own life or harm oneself. Similar to SA, any sentence with a mention of SI was labelled within the patient's notes. A SI annotation could be labeled as \textit{positive} or \textit{negative}, an example for each label is shown in Fig.~\ref{fig:pos_neg_si}. 

A sentence without SA or SI annotation would be considered as a \textit{neutral-SA} or \textit{neutral-SI} sentence respectively. Sentence level annotations provide more visibility to a medical expert for the hospital-stay level annotation.

\subsection{Dataset statistics}
ScAN consists of $19,690$ unique evidence annotations for the suicide relevant sections of $12,759$ EHRs of $697$ patient hospital-stays. There are a total of $17,723$ annotations for SA events and $1,967$ annotations for SI events. The distribution for both SA and SI events is provided in Table~\ref{table:sa_si_distr}.


\begin{table}[!htbp]
\renewcommand{\arraystretch}{1.2}
\footnotesize
\centering
\begin{tabular}{lccc}
\toprule
\multirow{2}{*}{\textit{General Statistics}} & Patients & Hospital-stays & Notes    \\ 
 & $669$ & $697$ & $12,759$       \\ \midrule
\multirow{2}{*}{\textit{Suicide Attempt}} & Positive & Negative & Unsure    \\ 
 & $14,815$ & $170$ & $2,738$       \\ \midrule
\multirow{2}{*}{\textit{Suicide Ideation}}  & Positive & Negative  \\ 
                  &   $1,167$     &   $800$  &  \\ \bottomrule
\end{tabular}
\caption{Distribution of unique annotations at the patient, hospital-stay and notes level in ScAN.}
\label{table:sa_si_distr}
\vspace{-1em}
\end{table}

%% file: sections/methodology.tex
In this section, we introduce \textit{ScANER} (\textbf{S}ui\textbf{c}ide \textbf{A}ttempt and Ideatio\textbf{n} \textbf{E}vents \textbf{R}etreiver): a strong baseline model for our dataset. ScANER consists of two sub-modules: (1) An \textit{evidence retriever module} that extracts the evidences related to SA and SI and (2) A \textit{predictor module} that predicts SA or SI label for the patient's hospital-stay using the evidences extracted by the retriever module.

\begin{table}[!htbp]
\renewcommand{\arraystretch}{1.3}
\footnotesize
\centering
\begin{tabular}{lcccc}
\toprule
\textit{Evidence} & \multicolumn{2}{c}{\textit{Yes}}      & \multicolumn{2}{c}{\textit{No}}                                                                \\ \midrule
Train         & \multicolumn{2}{c}{9,880}     & \multicolumn{2}{c}{30,133}                                                                 \\ 
Validation         & \multicolumn{2}{c}{1,803}     & \multicolumn{2}{c}{4,864}                                                                 \\ 
Test         & \multicolumn{2}{c}{3,038}     & \multicolumn{2}{c}{7,836}                                                                 \\ \midrule
\textit{SA}       & \textit{Positive} & \textit{Negative} & \textit{Unsure} &  \textit{Neutral-SA}\\  \midrule
Train         &  7,597      &    136     & 1,607      &     30,673                                                              \\ 
Validation         & 1,474        & 36        & 216      &  4,941                                                                 \\ 
Test         & 2,433        & 20        & 431      &    7,990                                                               \\ \midrule 
\textit{SI}       & \textit{Positive}  & \textit{Negative}   &     \textit{Neutral-SI}   &                                                                   \\ \midrule
Train         & 928        & 654        & 38,431     &                                                           \\ 
Validation         & 153        & 107         & 6,407 &                                                                  \\
Test     & 331        & 189        &  10,354    &                                                             \\\bottomrule
\end{tabular}
\caption{Distribution of evidences at paragraph level in ScAN for train, validation and test sets. A paragraph was considered an \textit{evidence}, labeled as \textit{Yes}, if it had at least one sentence annotated as SA or SI. A \textit{No} evidence paragraph was either \textit{Neutral-SA} or \textit{Neutral-SI}.} 
\label{table:para_distr}
\vspace{-1em}
\end{table}
\subsection{Evidence Retriever}
\paragraph{Problem Formulation:} Given an input clinical note, the model extracts the evidences (one or more sentences) related to SA or SI (SA-SI) from the note. This is a binary classification problem where given a text snippet the model predicts whether it has an evidence for SA-SI or not. We learn this task at paragraph level where the input is a set of $20$ consecutive sentences because the local surrounding context provides additional important information \cite{yang2021context, rawat2019naranjo}. A paragraph was labeled as \textit{evidence} \textit{no}, if all the sentences in that paragraph are \textit{neutral-SA} and \textit{neutral-SI}. If there was at least one SA-SI sentence, it was considered an \textit{evidence} \textit{yes}. As the number of non-evidence sentences significantly outsized the evidence sentences, we decided to use an overlapping window of $5$ sentences between the paragraphs to build more evidence paragraphs. The distribution of the paragraphs, across all evidence, SA and SI labels for train, validation, and test set is provided in Table~\ref{table:para_distr}. We segregated the train and test set such that any patient observed by the \textit{retriever module} during training was not seen in the test set. This is important as there are patients who had multiple hospital-stays in ScAN.
\vspace{0.4em}\noindent\newline 
\textbf{Proposed Model:} Transformer \cite{vaswani2017attention} based language models \cite{devlin2018bert, liu2019roberta} have shown great performance for a broad range of NLP classification tasks. Hence, to extract the evidence paragraphs we trained a Ro\textsc{bert}a \cite{liu2019roberta} based model. It has been previously shown that the domain-adapted versions of the pre-trained language models, such as clinical\textsc{bert} \cite{alsentzer2019publicly} or Bio\textsc{bert} \cite{lee2020biobert}, work better than their base versions. So, we further pre-trained the Ro\textsc{bert}a-base model over the MIMIC dataset to create a clinical version of Ro\textsc{bert}a model, hereby referenced as medRo\textsc{bert}a. During our initial exploration, we experimented with clinical\textsc{BERT} and Bio\textsc{BERT} but found that medRo\textsc{BERT}a consistently outperformed both models. medRo\textsc{BERT}a achieved an overall F1-score of $0.88$ whereas both clinical\textsc{BERT} and Bio\textsc{BERT} achieved an overall F1-score of $0.85$. Our hospital-level SA and SI predictor would work with any encoder-based evidence retriever model. 
\vspace{0.4em}\noindent\newline 
\textbf{Multi-task Learning:} We trained medRo\textsc{bert}a in a multi-task learning setting where along with learning the evidence classification task, the model also learns two auxiliary tasks: (a.) Identifying the label for SA between \textit{positive}, \textit{negative}, \textit{unsure} and \textit{neutral-SA} and, (b.) Identifying the label for SI between \textit{positive}, \textit{negative} and \textit{neutral-SI}. The training loss ($L(\theta)$) for our evidence retriever model was formulated as: 
\begin{equation}
    L(\theta) = L_{evi} + \alpha * L_{SA} + \beta * L_{SI}
\end{equation}

Where $L_{evi}$ is the negative log likelihood loss for evidence classification, $L_{SA}$ and $L_{SI}$ are SA and SI prediction losses respectively, and $\alpha$ and $\beta$ are the weights for the auxiliary tasks' losses. The distribution of labels across all the three tasks is highly skewed, hence, we applied the following techniques to learn an efficient and robust model. 
\begin{itemize}
    \item Weighted log loss was used in both main task and auxiliary tasks. The total loss for each task was calculated as the weighted sum of loss according to the \textit{label} of the input paragraph. Log weighing helps smooth the weights for highly unbalanced classes. The weight for each class was calculated using:
    \[
    w_{l,t}=\left\{
                \begin{array}{lll}
                  1.0 & \textrm{if}(w_{l,t}<1.0)\\
                  log(\gamma*N_t/N_{l,t})
                \end{array}
              \right.
    \]
    Where $N_t$ is the count of all training paragraphs for the task $t$ and $N_{l,t}$ is the count of paragraphs with label $l$ for the task $t$ and $w_{l,t}$ is the calculated weight for those paragraphs. We tuned $\gamma$ as a hyper-parameter. All training hyper-parameters for our best model are provided in Appendix~\ref{app:training_params}. \vspace{-0.4em}
    \item We also employed different sampling techniques \cite{youssef1999image}, up and down sampling, to help our model learn from an imbalanced dataset. After sweeping for different sampling combinations as hyper-parameters, we found that down-sampling the \textit{no}-evidence paragraphs by $10\%$ resulted in the best performance. \vspace{-0.4em}
    \item The \textit{negative} label of SA is severely under-represented in ScAN making it difficult for the model to learn useful patterns from such instances, refer Table~\ref{table:para_distr}. After discussion with the experts, we decided to group the instances of \textit{negative} and \textit{unsure} together and label them as \textit{neg\_unsure} because for both groups the general psych outcome is to let the patient leave after the hospital-stay as there is no solid evidence defining whether the self-harm was a SA event.
    
    
\end{itemize}

\begin{figure*}[!t]
    \centering
    \includegraphics[width=15cm]{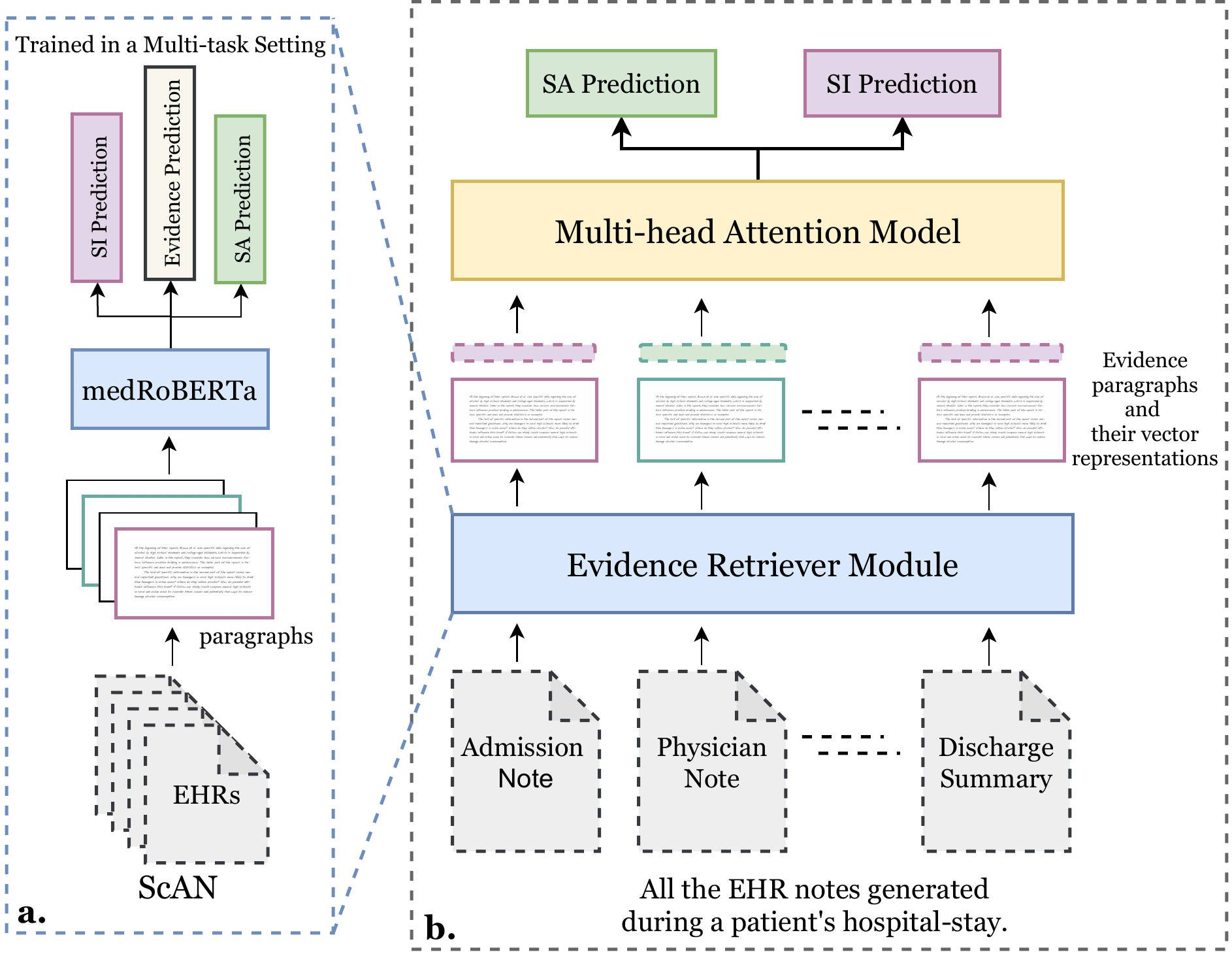}
    \caption{ScANER (\textbf{S}ui\textbf{c}ide \textbf{A}ttempt and Ideatio\textbf{n} \textbf{E}vents \textbf{R}etreiver) consists of two sub-modules: (a.) Evidence retriever module extracts evidence paragraphs from all EHR notes. We trained the module using all annotated paragraphs from ScAN. (b.) Prediction module predicts the SA and SI label for a patient using the evidence paragraphs extracted by the \textit{retriever module} from EHR notes during the patient's hospital-stay.}
    \label{fig:scaner}
\vspace{-1em}
\end{figure*}

\begin{table}[t]
\renewcommand{\arraystretch}{1.2}
\footnotesize
\centering
\begin{tabular}{lccc}
\toprule
\textit{Suicide Attempt} & Positive & Neg\_Unsure & Neutral-SA    \\ \midrule
Train & $377$ & $54$ &  $1,381$       \\ 
Val & $50$ & $10$ & $189$       \\ 
Test & $91$ & $19$ &  $326$      \\ \midrule
\textit{Suicide Ideation} & Positive & Negative & Neutral-SI    \\ \midrule
Train & $377$ & $214$ & $1,521$        \\ 
Val & $45$ & $28$ & $208$       \\ 
Test & $44$ & $35$ & $357$       \\ \bottomrule
\end{tabular}
\caption{Distribution of SA and SI at hospital-stay level in training, validation and testing set.}
\label{table:adm_distr}
\vspace{-2em}
\end{table}

\subsection{Hospital-stay level SA and SI Predictor}
\label{sec:adm_prediction}
\paragraph{Problem Formulation} Given all the clinical notes of a patient during the the hospital stay, the model predicts the label for SA (\textit{positive}, \textit{neg\_unsure} and \textit{neutral-SA}) and SI (\textit{positive}, \textit{negative} and \textit{neutral-SI}). The \textit{prediction module} uses the evidence paragraphs extracted by the \textit{retriever module}. 

\paragraph{Robust Finetuning} The retriever module is not perfect and can extract false positives. This results in extracting irrelevant paragraphs, with evidence label \textit{No}, along with evidence paragraphs for a hospital-stay with SA or SI and extracting irrelevant paragraphs as evidences for a hospital-stay with both SA and SI marked as \textit{neutral}. To tackle such situations and train a robust model, we applied three techniques: 
\vspace{-0.5em}
\begin{itemize}
    \item For a hospital-stay with a non-\textit{neutral} label for SA or SI, during training we added some noise in the form of irrelevant paragraphs (a paragraph with no SA or SI annotation) from the notes to the set of actual evidence paragraphs for the input. An irrelevant paragraph from a clinical note was sampled with a probability of $0.05$. This forced the predictor module to learn effectively even with noisy inputs. \vspace{-0.5em}
    \item For a \textit{neutral} hospital-stay with no evidence paragraphs, we randomly chose $X$ unique irrelevant paragraphs from the notes. $X$ was sampled from the distribution of number of evidence paragraphs of the non-\textit{neutral} hospital-stays. This prevented the leaking of any information to the prediction module during training by keeping the distribution of number of input paragraphs the same across \textit{neutral} and non-\textit{neutral} instances. \vspace{-0.5em}
    \item Since these hospital-stays were extracted using the ICD codes related to suicide and overdose, the data is quite skewed with only $102$ \textit{neutral} events from a total of $697$ hospital-stays. Whereas in a real-world scenario, \textit{neutral} hospital-stays would be much higher than non-\textit{neutral} ones. Hence, to facilitate a balanced learning of the predictor module we introduced $1,800$ \textit{neutral} hospital-stays from the MIMIC dataset. The distribution for SA and SI at hospital-stay level is provided in Table~\ref{table:adm_distr}. 
\end{itemize}

\begin{table*}[t]
\centering
\renewcommand{\arraystretch}{1.2}
\footnotesize
\begin{tabular}{l|ccc|l|ccc|l|ccc}
\toprule
\multicolumn{4}{c}{\textit{\textbf{Paragraph Evidence Prediction}}} & \multicolumn{4}{|c|}{\textit{\textbf{Paragraph SA Prediction}}} & \multicolumn{4}{c}{\textit{\textbf{Paragraph SI Prediction}}} \\ \midrule
    Evidence        & P       & R      & F      &          Labels       & P        & R       & F       &        Labels      & P         & R         & F        \\ \midrule
Yes    & 0.79    & 0.87   & 0.83   & Positive        & 0.71     & 0.74    & 0.73    & Positive      & 0.46      & 0.62      & 0.53     \\
No     & 0.95    & 0.91   & 0.93   & Neg\_Unsure     & 0.19     & 0.26    & 0.22    & Negative       & 0.38      & 0.46      & 0.42     \\

     \multicolumn{1}{c|}{-}    &     -    &      -  &    -    & Neutral-SA         & 0.95     & 0.92    & 0.93    & Neutral-SI      & 0.98      & 0.99      & 0.98     \\ \midrule
Overall     & 0.87    & 0.89   & 0.88   & Overall         & 0.62     & 0.64    & 0.63    & Overall      & 0.61      & 0.69      & 0.64    \\ \bottomrule 
\multicolumn{12}{r}{\scriptsize{P: Precision, R: Recall and F: F1-score.}}
\end{tabular}
\caption{Paragraph level performance of the \textit{evidence retriever module}. The overall evaluation metrics (precision, recall and F1-score) are macro-weighted. Evidence prediction is the main task whereas SA and SI prediction are auxiliary tasks and help the model align the vector representations of the paragraphs for the \textit{hospital-stay level suicidal behavior prediction}.}
\label{table:para_perf}
\vspace{-1em}
\end{table*}
\paragraph{Proposed Model}
The paragraphs extracted using the \textit{retriever module} for a patient's hospital-stay were provided as an input to the predictor module. We used a multi-head attention model to predict the SA and SI label for a hospital-stay as self-attention based models have proved to be quite effective for a lot of prediction tasks in machine learning \cite{devlin2018bert, cao2020self, hoogi2019self}. 

We encoded the extracted paragraphs ($[p_1, p_2 .... p_n]$) using the retriever module, medRo\textsc{BERT}a, to get a vector representation of $768$ dimensions for each of the paragraphs ($[v_1, v_2 ... v_n]$). Training the \textit{retriever module} on auxiliary tasks of predicting SA and SI helped align these paragraph representations for SA and SI prediction. Then, we added a prediction vector ($v_0$) along with all the vector representations of the paragraphs to get $\mathcal{V} = [v_0, v_1, v_2 ... v_n]$. We passed $\mathcal{V}$ through our multi-head attention model to get the hidden representations $\mathcal{H} = [h_0, h_1 ... h_n]$. We then passed $h_0$ through a SA inference layer and SI inference layer to predict the labels. During the whole training process, the weights of the retriever module were frozen whereas $v_0$ was a learnable vector initialised as an embedding in the multi-head attention model. We used a separate $v_0$ prediction vector so that it could retain the information from all the other paragraph representations for hospital-stay level prediction similar to how \textsc{[CLS]} is utilized in different transformer-based models for sequence prediction \cite{devlin2018bert, liu2019roberta}. We tuned the number of layers and number of attention heads of our prediction module as hyper-parameters and achieved the best performance using a 2-layer and 3-attention head model. Our complete ScANER model is illustrated in Fig~\ref{fig:scaner}.

%% file: sections/results.tex
Since the labels for both the retriever and prediction task are imbalanced, we used macro-weighted precision, recall, and F1-score to evaluate the overall performance of our models. Macro-weighted metrics provide better model insights across all labels.

\paragraph{Evidence Retriever Performance} Our multi-task learning model achieved a F1-score of $0.83$ for extracting positive evidence paragraphs and an F1-score of $0.88$ overall.
The retriever model has higher recall than precision for the positive evidence paragraphs ($0.87>0.79$), SA ($0.74>0.71$), and SI ($0.62>0.46$) events, as shown in Table~\ref{table:para_perf}. 
In healthcare, there is an incentive to maximize recall over precision \cite{watson2009systematic}. 
As mentioned in \S\ref{sec:adm_prediction}, ScANER was trained with added noisy paragraphs and is therefore robust to the extracted evidence paragraphs if they contain some false positives.

The \textit{retriever module} achieves an overall F1-score of $0.63$ for SA prediction and $0.64$ for SI prediction at paragraph-level. The performance for positive SA and SI evidence is much higher than the performance for \textit{neg\_unsure} SA and \textit{negative} SI. 
We looked at the confusion matrices for SA and SI paragraph-level prediction and found that largely ScANER made mistakes between \textit{positive} and \textit{neg\_unsure} labels for SA prediction and between \textit{positive} and \textit{negative} labels for SI prediction (refer Appendix~\ref{app:confusion_matrix}). The poor performance in SA for \textit{neg\_unsure} evidence prediction is mainly due to data sparsity where the \textit{neg\_unsure} cases are only $1743$; in contrast, the positive cases are 4-fold higher. Similarly, for SI the \textit{positive} cases are $1.4$ times higher than the \textit{negative} cases.


\begin{table}[t]
\footnotesize
\renewcommand{\arraystretch}{1.2}
\centering
\begin{tabular}{l|ccc}
\toprule
\multicolumn{4}{c}{\textit{\textbf{Hospital-stay SA Prediction}}}  \\ \midrule
Labels        & Precision  & Recall  & F1-score \\ \midrule
Positive      & 0.81       & 0.93    & 0.87     \\
Neg\_Unsure   & 0.48       & 0.58    & 0.52     \\
Neutral-SA       & 0.98       & 0.93    & 0.96     \\ \midrule
Overall       & 0.76       & 0.81    & 0.78     \\ \midrule
\multicolumn{4}{c}{\textit{\textbf{Hospital-stay SI Prediction}}} \\ \midrule
Labels        & Precision  & Recall  & F1-score \\ \midrule
Positive      & 0.49       & 0.93    & 0.65     \\
Negative   & 0.40        & 0.11    & 0.18     \\
Neutral-SI       & 0.99       & 0.95    & 0.97     \\ \midrule
Overall       & 0.63       & 0.66    & 0.60     \\ \bottomrule
\end{tabular}
\caption{Hospital-stay level SA and SI prediction performance of ScANER. }
\label{table:adm_perf}
\vspace{-1em}
\end{table}

\paragraph{Hospital-stay level Prediction Performance} Our multi-head attention model is able to achieve an overall macro F1-score of $0.78$ for SA prediction and $0.60$ for SI prediction, as shown in Table~\ref{table:adm_perf}. 
For SA, the \textit{prediction module} achieves a recall of $0.93$ for the positive label. After analysing the confusion matrix, the model largely predicts a \textit{positive} label for the visits with \textit{neg\_unsure} label, as shown in Table~\ref{table:confusion_matrix}. The poor performance for \textit{neg\_unsure} is largely because of its small representation in the training set of ScAN, $54$ negative cases as compared to $377$ positive and $1,381$ neutral instances. In our future work, we plan to expand ScAN with more instances of \textit{negative} SA events.

\begin{table}[!htbp]
\centering
\renewcommand{\arraystretch}{1.2}
\footnotesize
\begin{tabular}{l|ccc}
\toprule
\multicolumn{4}{c}{\textbf{\textit{Hospital-stay SA Prediction}}}            \\ \midrule
            & Positive & Neg\_Unsure & Neutral-SA \\  \midrule
Positive    & 85       & 4           & 2       \\
Neg\_Unsure & 5        & 11          & 3       \\
Neutral-SA     & 15       & 8           & 303    \\ \midrule
\multicolumn{4}{c}{\textbf{\textit{Hospital-stay SI Prediction}}}            \\ \midrule
        & Positive & Negative & Neutral-SI \\ \midrule
Positive & 41      & 2      & 1       \\
Negative  & 27      & 4      & 4       \\
Neutral-SI & 15      & 4      & 338    \\ \bottomrule

\end{tabular}
\caption{Confusion matrices for SA and SI prediction at hospital-stay level.}
\label{table:confusion_matrix}
\vspace{-2em}
\end{table}

\begin{figure}[!htbp]
    \centering
    \includegraphics[width=7.7cm]{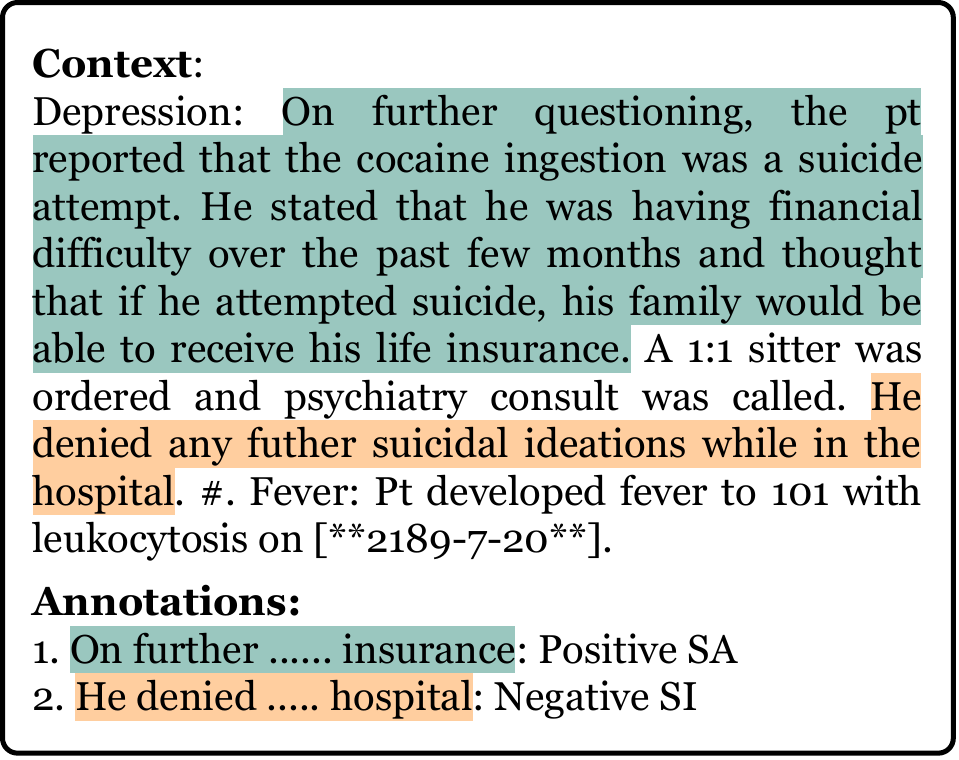}
    \caption{An instance for which ScANER incorrectly predicted a \textit{negative} hospital-level SI as \textit{positive}.}
    \label{fig:si_error}
    \vspace{-1.0em}
\end{figure}

For SI, the prediction module achieves an overall F1-score of $0.60$ with a precision of $0.63$ and recall of $0.66$. The model has a high recall for \textit{neutral-SI} and \textit{positive} but the \textit{positive} label has a low precision of $0.49$. After analysing the test set, we observed that a lot of patient hospital-stays with \textit{negative} labels are getting wrongly predicted as \textit{positive}, as shown in Table~\ref{table:confusion_matrix}. After doing error analysis for hospital-stays with \textit{negative} labels, we observed that a lot of extracted evidence paragraphs contain information that suggests that the patient had SI before the SA but does not have SI anymore during the hospital-stay. As shown in the example in Fig~\ref{fig:si_error}, the past SI is an explanation for the SA but then the patient does not have any further SI during the hospital-stay. This suggests that period assertions for these annotations are quite important and we aim to add period assertion property in our future work by further annotating ScAN.

%% file: sections/appendix.tex

\section{Selected Clinical Sections}
\label{app:clinical_sections}
The sections selected for annotations after using \texttt{clinical\_sectionizer} are enumerated below:
\begin{enumerate}
\item Allergy
\item Case Management 
\item Consult
\item Discharge Summary
\item Family history
\item General
\item HIV Screening
\item Labs and Studies
\item Medication
\item Nursing
\item Nursing/other
\item Nutrition
\item Observation and Plan
\item Past Medical History
\item Patient Instructions
\item Physical Exam
\item Physician 
\item Present Illness
\item Problem List
\item Radiology
\item Rehab Services
\item Respiratory 
\item Sexual and Social History
\item Social Work
\end{enumerate}

\section{Hyper-parameter Settings}
\label{app:training_params}
All the hyper-parameter settings for both modules of ScANER are provided in Table~\ref{table:hyper_params}.
\begin{table}[!htbp]
\centering
\footnotesize
\renewcommand{\arraystretch}{1.2}
\begin{tabular}{cccc}
\toprule
\multicolumn{4}{c}{\textbf{\textit{Evidence Retriever Module}}}                        \\ \midrule
Learning Rate             & Warmup steps                 & Optimizer            & Adam $\epsilon$ \\
2e-5                  & 2,000                        & Adam                 & 1e-8                     \\ \midrule
$\gamma$     & $\alpha$        & $\beta$ &                              \\
2.5                       & 1.1                          & 1.5                  &                              \\ \midrule
\multicolumn{4}{c}{\textbf{\textit{Hospital Stay SA-SI Prediction Module}}}                        \\ \midrule
Attention Heads           & Attention Layers             & Learning Rate        & Warmup steps                 \\
3                         & 2                            & 1e-3             & 1,200                        \\ \midrule
Optimizer                 & Adam $\epsilon$ &                      &                              \\ 
Adam                      & 1e-8                     &                      &                             \\ \bottomrule
\end{tabular}
\caption{Hyper-parameter setting for both retriever and prediction module of ScANER.}
\label{table:hyper_params}
\end{table}
\newpage
\section{Confusion matrices}
\label{app:confusion_matrix}
The confusion matrices for SA and SI prediction at paragraph level is provided in Table~\ref{table:conf_app}.
\begin{table}[!htbp]
\renewcommand{\arraystretch}{1.2}
\footnotesize
\centering
\begin{tabular}{l|ccc}
\toprule
\multicolumn{4}{c}{\textbf{\textit{Paragraph SA Prediction}}}            \\ \midrule
            & Positive & Neg\_Unsure & Neutral-SA \\ \midrule
Positive    & 1,804    & 285         & 344     \\
Neg\_Unsure & 253      & 118         & 80      \\
Neutral-SA     & 472      & 204         & 7,314   \\ \midrule
\multicolumn{4}{c}{\textit{\textbf{Paragraph SI Prediction}}}           \\ \midrule
            & Positive   & Negative      & Neutral-SI \\ \midrule
Positive     & 206      & 69          & 56      \\
Negative      & 71       & 87          & 31      \\
Neutral-SI     & 170      & 73          & 10,111 \\ \bottomrule
\end{tabular}
\caption{Confusion matrices for the predictions on the test set of evidence retriever.}
\label{table:conf_app}
\end{table}